\documentclass[conference]{IEEEtran}
\IEEEoverridecommandlockouts

\usepackage{cite}
\usepackage{amsmath,amssymb,amsfonts}
\usepackage{algorithmic}
\usepackage{graphicx}
\usepackage{textcomp}
\usepackage{xcolor}

\usepackage{multirow}
\usepackage{booktabs}
\usepackage{tabularx}

\usepackage{tikz, hyperref}
\newcommand\copyrighttext{%
\footnotesize \textcopyright 2026 IEEE. Personal use of this material is permitted. Permission from IEEE must be obtained for all other uses, in any current or future media, including reprinting/republishing this material for advertising or promotional purposes, creating new collective works, for resale or redistribution to servers or lists, or reuse of any copyrighted component of this work in other works.
}
\newcommand\copyrightnotice{%
\begin{tikzpicture}[remember picture,overlay]
\node[anchor=south,yshift=10pt] at (current page.south) {\fbox{\parbox{\dimexpr\textwidth-\fboxsep-\fboxrule\relax}{\copyrighttext}}};
\end{tikzpicture}%
}

\def\BibTeX{{\rm B\kern-.05em{\sc i\kern-.025em b}\kern-.08em
    T\kern-.1667em\lower.7ex\hbox{E}\kern-.125emX}}
\begin{document}

\title{BehaviorFormer: Discovering Behavioral Patterns via Clusterable Next-Token Prediction\\
\thanks{This research has been funded by the Federal Ministry of Education and Research of Germany and the state of North-Rhine Westphalia as part of the Lamarr-Institute for Machine Learning and Artificial Intelligence.}
}

\author{\IEEEauthorblockN{Tobias Schneider\textsuperscript{1,2},
Danial Fahrner\textsuperscript{1},
Soren Antebi\textsuperscript{1},
Daniel Uedelhoven\textsuperscript{1},\\
Lennard Bodden\textsuperscript{1},
Sandra Halscheidt\textsuperscript{1},
Rafet Sifa\textsuperscript{1,2}}
\IEEEauthorblockA{\textsuperscript{1}\textit{Fraunhofer IAIS}, Sankt Augustin, Germany \\
\textsuperscript{2}\textit{University of Bonn}, Bonn, Germany \\
}
}

\maketitle
\copyrightnotice

\IEEEpubidadjcol

\begin{abstract}
Discovering behavioral patterns in open-ended games is challenging due to the lack of clear objectives or ground-truth annotations.
In this work, we extend a next-token prediction Transformer with the deep clustering algorithm DCSS to learn meaningful representations with well-separated clusters.
In a controlled Minecraft session with 26 players, the method discovers 23 distinct behavioral patterns that generalize across the studied players, further refining structures present in self-supervised representations into clearly separable clusters.
We group the clusters into seven interpretable categories covering stationary, roaming, travel, mining, base work, logging, and farming behavior.
\end{abstract}


\section{Introduction}
Player behavioral modeling is challenging, especially in open-ended games such as Minecraft, where players follow no clear goal.
Unlike objective-driven games, where behavior is primarily shaped by the game objective, open-ended games encourage players to play however they want, leaving no ground truth for what counts as a behavioral pattern.
This makes manual analysis based on hand-crafted features and self-reported annotations challenging.
It requires defining behavioral patterns upfront, embeds strong prior assumptions, and struggles to capture unexpected behavior that does not fit into the predefined taxonomy.
Self-reported annotations are subjective and expensive to collect, while hand-crafted features fail to represent the complex temporal structure of player behavior.
This motivates automatic behavioral pattern discovery directly from raw behavior data using deep learning.
Mapping player behavior onto a small set of behavioral patterns compresses complex behavioral data into interpretable high-level behavior that both humans and algorithms can reason about.
This can be useful for various applications such as behavioral analytics, player modeling for personalization, content design and balancing, or churn and engagement analysis.

Self-supervised sequence models such as Transformers~\cite{transformer} can capture temporal dynamics but do not explicitly model behavioral patterns.
We address this by combining a Transformer with the Deep Clustering using Soft Silhouette (DCSS)~\cite{dcss} algorithm to jointly learn representations and compact, well-separated clusters corresponding to behavioral patterns.
We apply this approach to a controlled Minecraft session with 26 players, finding 23 global behavioral patterns that can be categorized into seven interpretable groups.
The discovered patterns generalize across players, capturing globally shared behavior rather than individual playstyles.

\section{Related Work}

Prior work on Minecraft player modeling uses supervised classification on manually extracted features and self-reported labels.
Müller et al.~\cite{minecraft_behavior} distinguish between building, mining, fighting, and exploring behavior with 65\% accuracy.
While this shows that high-level behavioral patterns can be derived from raw game logs to some extent, the approach is limited for behavioral discovery as categories and features must be defined in advance.
The moderate performance suggests that hand-crafted representations only partially capture the ambiguity and temporal structure of open-ended player behavior.

Similarly, prior work on player profiling with traditional machine learning algorithms relies on hand-crafted features~\cite{drachen_clustering,sifa_behavior,sifa_profiling}.
This work shows that clustering can produce actionable player profiles, but the resulting structure still depends on feature engineering and game-specific design knowledge.
Player2Vec~\cite{player2vec} reduces this dependency by treating game event logs as token sequences and training a Transformer with a next-token prediction (NTP) objective to learn player representations without explicit behavior labels.
Pu et al.~\cite{mlm_repr} take a similar approach using masked language modeling.
However, since the objectives are primarily designed for sequence modeling, cluster structure emerges only implicitly.

Deep clustering methods instead learn representations and clusters jointly.
DEC~\cite{dec} and IDEC~\cite{idec} encourage compact clusters through a KL-divergence objective, but do not explicitly optimize inter-cluster separation.
DCSS~\cite{dcss} addresses this with a differentiable soft silhouette objective that rewards both compactness and separation of clusters, making it a promising method for behavioral pattern discovery.

\section{Method}

\subsection{Clustering Objective}

The DCSS objective combines a reconstruction loss with a soft silhouette clustering loss and an entropy regularizer to prevent trivial solutions:
\begin{equation}
\label{eq:dcss_loss}
\mathcal{L} = \mathcal{L}_{rec} + \lambda_1 \bigl(1 - Sf(h_r(X))\bigr) - \lambda_2 \bar{H}(h_r(X)),
\end{equation}
where $\mathcal{L}_{rec}$ is the chosen reconstruction loss, $Sf(\cdot)$ is the soft silhouette score over the cluster assignment probabilities $h_r(X)$, $\bar{H}(\cdot)$ is the mean entropy of the assignment probabilities, and $\lambda_1, \lambda_2$ balance the three terms.

In this work, we replace DCSS's autoencoder with a Transformer for its sequence modeling capabilities, using the final token's hidden state as input to the clustering network $h_r$.
As our inputs are discrete tokens, we use NTP cross-entropy as $\mathcal{L}_{rec}$ instead of MSE.

\subsection{Dataset Construction}
We collected data from one session with 26 individuals playing the game Minecraft over 6 hours.
To motivate diverse behavior, the players were not given any concrete goal and had complete freedom in what to do during the session, except to form groups of 2-6 players.
The session was recorded in a controlled environment, allowing verification of qualitative analysis, while maintaining rich behavioral patterns.
The following game events were logged: Breaking or placing a block, picking up or crafting an item, interacting with a chest or furnace, and death of an entity such as an enemy monster.
We also collected the position of each player at an interval of 0.5 seconds, further contextualizing the logged events.

To construct a dataset for NTP using Transformers, we tokenized the collected logs. 
For the event logs, each unique event is assigned an action token.
For the movement logs, both direction and magnitude are discretized: direction into 8 directional bins, and magnitude into thresholds suited to Minecraft, differentiating between standing still, sneaking, walking, sprinting, and very fast motion when using a boat.
Each possible combination is assigned one unique motion token.
The resulting vocabulary contains 532 tokens.
We split the data into sequences of 64 tokens, corresponding to approximately 30 seconds of gameplay.
This length was chosen in preliminary experiments, as shorter sequences fail to capture meaningful behavioral patterns while longer sequences mix multiple patterns per sequence.
In total, we collected 11,234 sequences across 26 players.
For evaluation of NTP loss on unseen data, we hold out 60 randomly sampled sequences per player as a validation set in each run, leaving 9,674 sequences for training.

We provide the Transformer with two learned additive positional embeddings:
a token position embedding $T_{pos}$ based on the token's index in the sequence,
and a tick position embedding $T_{tick}$ shared across all tokens within the same 0.5\,s window, capturing events that occur simultaneously.

\subsection{Hyperparameter Selection}

In preliminary experiments, we identified parameters with significant impact on performance and swept them over an extensive hyperparameter grid as shown in Table~\ref{tab:ablation_abs}.
For the Transformer, we vary embedding dimension $d$, depth $L$, and DCSS-stage learning rate $\eta$.
For DCSS, we vary the number of clusters $K$, clustering loss weight $\lambda_1$, pretraining epochs $N_\text{pre}$, and clustering temperature $\tau_c$.
We train with a batch size of 256 and a pretraining learning rate of $10^{-4}$, use the default $\lambda_2=0.01$ from the original DCSS paper, 2 attention heads, no dropout, and early stopping with a patience of 10 epochs on validation loss.

To identify a robust selection of hyperparameters that consistently leads to good clustering, we focus on four key metrics aggregated across ten experiments: %
\begin{enumerate}
    \item Silhouette score: measures how compact and separated clusters are, which we interpret as a measure of cluster \textbf{quality}.
    \item Adjusted Rand Index (ARI): quantifies agreement between two clusterings, adjusted for chance. We report the mean pairwise ARI across 10 independent runs as a measure of cluster assignment \textbf{stability}.
    \item Cluster utilization: quantifies the ratio of clusters with assigned samples, which we interpret as \textbf{coverage}.
    \item Loss: measures next-token prediction performance using cross-entropy on the validation set, which we interpret as representation \textbf{fidelity}.
\end{enumerate}

\section{Results \& Discussion}

\begin{figure*}
    \centering
    \includegraphics[width=1.0\textwidth,height=2.5in]{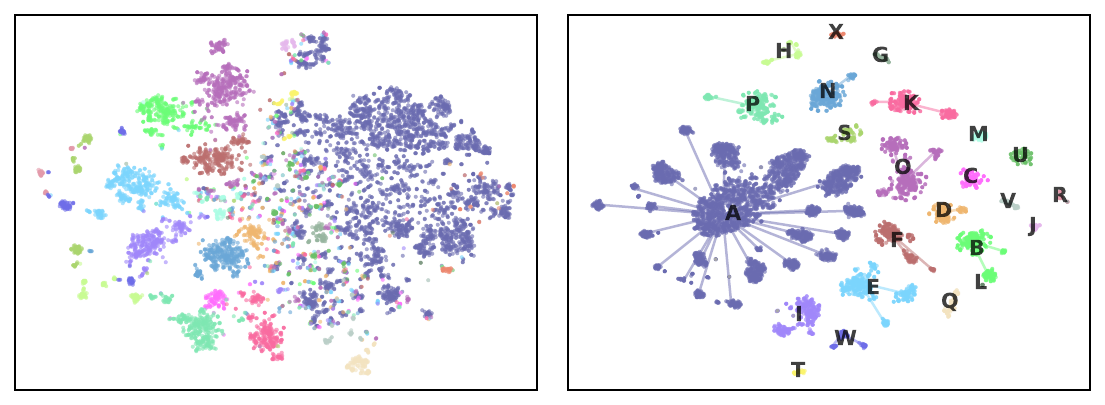}
    \caption{T-SNE visualization of embedding spaces learned by NTP alone (left) and DCSS (right), colored by DCSS cluster assignment. In the DCSS space, sub-clusters detected via HDBSCAN are indicated by connecting lines within their parent cluster.}
    \label{fig:tsne}
\end{figure*}

\subsection{Hyperparameter Analysis}

No configuration is optimal across all four objectives, as they exhibit clear trade-offs (e.g. cluster quality vs. representation fidelity).
However, with multi-objective optimization we selected a robust configuration that achieves near-optimal values on all objectives.
Table~\ref{tab:ablation_abs} depicts the impact of modulating each hyperparameter one at a time on each metric, using the selected configuration as baseline.

\begin{table}[h]
\centering
\caption{Local ablation study for selected hyperparameters. The selected parameter value and the best value for each metric are shown in bold.}
\label{tab:ablation_abs}
\begin{tabular*}{\columnwidth}{@{}c@{\hspace{5pt}}l@{\extracolsep{\fill}}cccc@{}}
  \toprule
  \multicolumn{2}{c}{\textbf{Parameter}}
      & \textbf{Silhouette}$\uparrow$
      & \textbf{ARI}$\uparrow$
      & \textbf{Loss}$\downarrow$
      & \textbf{Utilization}$\uparrow$ \\
  \midrule
   & 10 & $\mathbf{99.7}{\pm}0.3$ & $68.4{\pm}33.4$ & $1.259{\pm}.006$ & $92.0{\pm}22.0$ \\
  $K$ & \textbf{25} & $99.5{\pm}0.0$ & $\mathbf{98.2}{\pm}0.7$ & $\mathbf{1.255}{\pm}.003$ & $\mathbf{95.6}{\pm}3.0$ \\
   & 50 & $99.3{\pm}0.0$ & $95.8{\pm}4.6$ & $1.255{\pm}.002$ & $85.4{\pm}3.5$ \\
  \midrule
   & \textbf{64} & $\mathbf{99.5}{\pm}0.0$ & $\mathbf{98.2}{\pm}0.7$ & $\mathbf{1.255}{\pm}.003$ & $\mathbf{95.6}{\pm}3.0$ \\
  $d$ & 128 & $99.4{\pm}0.1$ & $55.9{\pm}40.7$ & $1.271{\pm}.013$ & $71.6{\pm}32.2$ \\
   & 256 & $98.8{\pm}0.7$ & $64.3{\pm}45.0$ & $1.292{\pm}.013$ & $75.2{\pm}25.3$ \\
  \midrule
   & 1 & $99.2{\pm}0.1$ & $88.5{\pm}7.5$ & $1.268{\pm}.002$ & $\mathbf{96.0}{\pm}1.9$ \\
  $L$ & 2 & $99.5{\pm}0.0$ & $\mathbf{98.3}{\pm}0.5$ & $1.261{\pm}.003$ & $92.0{\pm}3.3$ \\
   & \textbf{3} & $\mathbf{99.5}{\pm}0.0$ & $98.2{\pm}0.7$ & $\mathbf{1.255}{\pm}.003$ & $95.6{\pm}3.0$ \\
  \midrule
   & 1e-4 & $99.5{\pm}0.2$ & $83.1{\pm}16.8$ & $1.261{\pm}.002$ & $94.4{\pm}2.8$ \\
  $\eta$ & \textbf{5e-4} & $99.5{\pm}0.0$ & $\mathbf{98.2}{\pm}0.7$ & $\mathbf{1.255}{\pm}.003$ & $\mathbf{95.6}{\pm}3.0$ \\
   & 1e-3 & $\mathbf{99.7}{\pm}0.2$ & $34.1{\pm}44.3$ & $1.258{\pm}.004$ & $57.6{\pm}41.1$ \\
  \midrule
   & 10 & $99.5{\pm}0.1$ & $68.8{\pm}19.3$ & $1.261{\pm}.004$ & $83.2{\pm}13.2$ \\
  $N_\text{pre}$ & \textbf{25} & $\mathbf{99.5}{\pm}0.0$ & $98.2{\pm}0.7$ & $\mathbf{1.255}{\pm}.003$ & $\mathbf{95.6}{\pm}3.0$ \\
   & 50 & $99.3{\pm}0.1$ & $\mathbf{98.4}{\pm}1.1$ & $1.256{\pm}.002$ & $94.8{\pm}5.7$ \\
  \midrule
  \multirow{4}{*}{$\lambda_1$} & 0.01 & $54.3{\pm}4.7$ & $77.5{\pm}10.5$ & $\mathbf{1.252}{\pm}.003$ & $\mathbf{97.6}{\pm}2.8$ \\
   & 0.1 & $95.3{\pm}0.7$ & $87.1{\pm}17.0$ & $1.256{\pm}.002$ & $94.0{\pm}4.7$ \\
   & 0.5 & $99.2{\pm}0.0$ & $\mathbf{98.5}{\pm}0.4$ & $1.255{\pm}.002$ & $93.6{\pm}2.8$ \\
   & \textbf{1.0} & $\mathbf{99.5}{\pm}0.0$ & $98.2{\pm}0.7$ & $1.255{\pm}.003$ & $95.6{\pm}3.0$ \\
  \midrule
   & 5 & $99.5{\pm}0.2$ & $75.0{\pm}37.8$ & $1.257{\pm}.004$ & $82.4{\pm}29.1$ \\
  $\tau_c$ & \textbf{10} & $99.5{\pm}0.0$ & $98.2{\pm}0.7$ & $\mathbf{1.255}{\pm}.003$ & $\mathbf{95.6}{\pm}3.0$ \\
   & 20 & $\mathbf{99.6}{\pm}0.0$ & $\mathbf{98.7}{\pm}0.3$ & $1.256{\pm}.002$ & $86.8{\pm}4.6$ \\
  \bottomrule
\end{tabular*}
\end{table}

Both stability and coverage of clustering seem to heavily depend on the number of clusters $K$.
The selected configuration does not achieve optimal coverage, meaning that only 24 of the 25 clusters are used.
A small embedding dimension $d$ of 64 is crucial for good performance across all metrics. 
The method seems to benefit from deeper models, as higher $L$ improves fidelity while maintaining stability and coverage.
The model is sensitive to the learning rate $\eta$, as both lower and higher learning rates show instability and inconsistent coverage.
Longer pretraining improves clustering stability and coverage.

While the DCSS paper uses a $\lambda_{1}$ of 0.01, we found that much higher weighting for the soft silhouette loss was required to achieve good separability and stability, which might originate from significant differences in the dataset and model architecture.
Also, we observed that the clustering softmax temperature $\tau_c$ can benefit from slightly lower values to prevent low cluster utilization, while the DCSS paper uses a value of 20.

Our results show that the method is sensitive to the selected hyperparameters, making their selection crucial.
The selected configuration achieves highly separable clusters with high stability and good coverage. 
The loss only increases by $1\%$ over using raw NTP, indicating that the clustering does not hurt NTP capabilities significantly, which suggests meaningful clustering, but requires further qualitative analysis.

Figure~\ref{fig:tsne} shows t-SNE~\cite{tsne} visualizations of the embedding spaces learned by NTP alone and DCSS.
The samples are colored by DCSS cluster assignment to allow comparison of global structure across the methods.
The NTP space already exhibits cluster-like structures, but applying clustering algorithms directly to it does not yield satisfactory results, as many structures lack clear boundaries.
In contrast, the DCSS space shows 24 well-separated clusters of various sizes, with some exhibiting sub-cluster structures detected via HDBSCAN~\cite{hdbscan}.
The shared coloring reveals that most clusters extracted by DCSS correspond to structures already present in the NTP space, but with noisy boundaries and overlap between neighboring groups.
This confirms that DCSS refines latent structure already present in the NTP representations into clearly separable clusters, with negligible impact on representation fidelity.

\subsection{Qualitative Analysis}

\begin{table*}[t]
\centering
\caption{Discovered behavioral clusters, grouped into seven interpretable categories. Letters in parentheses reference consistent IDs.}
\label{tab:clusters}
\begin{tabularx}{\textwidth}{l X l r}
\toprule
\textbf{Group} & \textbf{Clusters} & \textbf{Description} & \textbf{N} \\
\midrule
Stationary & Stationary (A) & Idle or low-activity without movement & 6318 \\
Roaming & Relocating (B), Wandering (C), Crop tending (D), Mixed (E), Resource gathering (F), Aquatic foraging (G), Wood gathering (H), Provisioning (I) & Local movement with some opportunistic activity & 3392 \\
Travel & Short (P), Moderate (L), Long (M), Boat (T) & Directed long-range movement & 486 \\
Base work & Crafting (U), Building (O), Remodeling (K), Earthwork (S) & Construction and item production at bases & 424 \\
Mining & Tunneling (N), Caving (R), Excavating (J) & Underground resource extraction & 423 \\
Logging & Oak/Birch (V), Spruce (W) & Intensive wood logging & 105 \\
Farming & Farming (Q) & Planting and harvesting crops & 80 \\
\bottomrule
\end{tabularx}
\end{table*}
To interpret the clusters, we compute statistics over action and motion token distributions, player height, position displacement, distance to the nearest player base, player proximity, session timing, and inter-cluster transition probabilities.
We then manually assign post-hoc cluster labels in Table~\ref{tab:clusters} that reflect the dominant behavior, and merge them into seven interpretable categories with shared high-level behavior.
One of the 24 utilized clusters (X), with only 6 assigned samples, was excluded from the analysis due to insufficient size, resulting in a total of 23 clusters used for analysis.

The Stationary~(A) cluster accounts for over half of samples and exhibits an exceptionally high self-transition rate of 0.71, meaning players remain in this state across consecutive sequences far more often than in any other cluster.
Especially for this cluster, the analysis of sub-clusters was essential for validating cluster interpretations.
We found that the largest sub-cluster consists of pure idle behavior with neither actions nor movement, while smaller sub-clusters include some actions.
However, the diverse activities are unified into a single cluster because they are all dominated by low movement.
Roaming clusters account for the second largest portion, capturing players between focused and stationary work, performing opportunistic actions such as collecting resources while on the move.
These clusters have low self-transition rates below 0.1, meaning players typically stay briefly before going back to stationary patterns or into focused work.
Together, Stationary and Roaming account for over 85\% of samples, showing that intensive activity is comparatively rare.

Focused work clusters (Mining, Base work, Farming, and Logging) make up less than 10\% of the samples, but exhibit the highest activity with an average of 11 to 17 action tokens per sequence.
Mining clusters differ in extraction strategy: Tunneling~(N) captures systematic tunnel construction, while Caving~(R) reflects natural cave exploration, both with the goal of finding ores.
Excavating~(J) captures large-scale stone breaking without a clear structure.
These interpretations are confirmed by low player height well below surface level. 
Base work clusters capture near-base construction and maintenance.
Earthwork~(S) in particular appears predominantly in the first half of the session, which aligns with player base setup.
Farming~(Q) captures farm construction and crop harvesting, and is the most action-intensive cluster overall.
As each action is assigned a unique token, cutting down trees of types that grow in different regions ends up in different clusters (V, W).

Travel clusters form a progression from short overland sprints~(P) to steady boat travel~(T), and consist almost entirely of motion tokens.
They appear especially in the session's second half, capturing targeted travel and raiding that occurred later in the session.
Boat travel is the second most persistent state with a self-transition rate of 0.38, reflecting sustained ocean crossings across many consecutive sequences.

Nearly half of all clusters cover at least 24 of 26 players, and every cluster covers at least 10, confirming that the discovered clusters capture shared behavior rather than individual playstyles.
Common patterns like Stationary and Roaming span nearly all players, while specialized ones such as Farming, Logging, or Boat travel cover fewer players, as expected since not every player engages in every activity.
Although combat clusters are missing due to the relevant game events not being logged, the extracted patterns align with behavior observed during the controlled session, confirming that the method recovers meaningful behavioral structure without requiring self-reported annotations.

\section{Conclusion}
In this work, we combined a Transformer with the DCSS clustering objective to enable automatic behavioral pattern discovery from raw gameplay logs in open-ended games.
We found that careful hyperparameter selection is crucial for achieving high-quality and stable results.
Applied to a controlled Minecraft session, the method yields 23 well-separated clusters that group into seven interpretable behavioral categories.
The discovered patterns are global, capturing shared behavior across players rather than individual playstyles, and correspond to refinements of structures already present in pure NTP representations.
These results motivate further work to validate generalization across games and longer sessions, investigate the impact of sequence construction, tokenization, and architecture choices, and explore methods for dynamically determining the number of clusters.

\bibliographystyle{IEEEtran}
\bibliography{IEEEabrv,references}

\end{document}